# Internal-transfer Weighting of Multi-task Learning for Lung Cancer Detection

Yiyuan Yang[a], Riqiang Gao*[a], Yucheng Tang[b], Sanja L. Antic[c], Steve Deppen[c], Yuankai Huo[a],
Kim L. Sandler[c], Pierre P. Massion[c], Bennett A. Landman[a,b]

[a] Computer Science, Vanderbilt University, Nashville, TN, USA 37235
[b] Electrical Engineering, Vanderbilt University, Nashville, TN, USA 37235
[c] Vanderbilt University Medical Center, Nashville, TN, USA 37235
(*Corresponding Author: riqiang.gao@vanderbilt.edu)

## ABSTRACT

Deep learning has achieved many successes in medical imaging, including lung nodule segmentation and lung cancer prediction on computed tomography (CT). Recently, multi-task networks have shown to both offer additional estimation capabilities, and, perhaps more importantly, increased performance over single-task networks on a "main/primary" task. However, balancing the optimization criteria of multi-task networks across different tasks is an area of active exploration. Here, we extend a previously proposed 3D attention-based network with four additional multi-task subnetworks for the detection of lung cancer and four auxiliary tasks (diagnosis of asthma, chronic bronchitis, chronic obstructive pulmonary disease, and emphysema). We introduce and evaluate a learning policy, Periodic Focusing Learning Policy (PFLP), that alternates the dominance of tasks throughout the training. To improve performance on the primary task, we propose an Internal-Transfer Weighting (ITW) strategy to suppress the loss functions on auxiliary tasks for the final stages of training. To evaluate this approach, we examined 3386 patients (single scan per patient) from the National Lung Screening Trial (NLST) and de-identified data from the Vanderbilt Lung Screening Program, with a 2517/277/592 (scans) split for training, validation, and testing. Baseline networks include a single-task strategy and a multi-task strategy without adaptive weights (PFLP/ITW), while primary experiments are multi-task trials with either PFLP or ITW or both. On the test set for lung cancer prediction, the baseline single-task network achieved prediction AUC of 0.8080 and multi-task baseline failed to converge (AUC 0.6720). However, applying PFLP helped multi-task network clarify and achieved test set lung cancer prediction AUC of 0.8402. Furthermore, our ITW technique boosted the PFLP enabled multi-task network and achieved an AUC of 0.8462 (McNemar test, $p < 0.01$). In conclusion, adaptive consideration of multi-task learning weights is important, and PFLP and ITW are promising strategies.

## 1. INTRODUCTION

According to the World Health Organization's 2018 September report, there were 2.09 million new lung cancer cases and 1.75 million deaths caused by lung cancer in 2018 alone, making lung cancer one of the most common and deadliest form of cancer [8]. Recently, the U.S. Preventive Services Task Force (USPSTF) recommended lung screening with computed tomography (CT) for older individuals with a substantial history of smoking [9]. With the advancement of deep learning and big data, lung cancer detection has received intense attention in both academia and industry. In the past decade, a variety of deep learning methods have been proposed for lung cancer detection, including both single-task and multi-task learning. Single-task learning saw progresses in both nodule mask assisted training [4][5] and longitudinal analysis [10][11] and multi-task learning also proved to be promising. Many previous multi-task studies [1][2][5][7][12] selected nodule detection, segmentation, and survival analysis as the auxiliary tasks that accompany lung cancer diagnosis. Meanwhile, other multi-task studies performed learning on clinical data as complementary learning for imaging data [6]. However, such approaches used traditional probabilistic learning rather than deep learning. In this study, we proposed a new multi-task learning method to improve the lung cancer diagnosis task as the central task, with diagnoses of four common pulmonary airway diseases (i.e., adult asthma, chronic bronchitis, chronic obstructive pulmonary disease, and emphysema) as auxiliary tasks in a single end-to-end convolutional neural network. Success in multi-task learning with low-dose CT could significantly broaden how CT is used as a screening modality.

In the proposed network, the five tasks above share the same encoder of the network but have their own task-specified fully connected blocks. In our experiments, the total loss is the weighted sum of the losses using our empirically changing inter-class loss weights. Conventional multi-task learning calculates the final loss to be sum of the tasks' losses or a weighted sum of the losses using the same inter-class loss weights list throughout training. This is analogous to a simultaneously tracking four conversations in a room of people speaking simultaneously. To address this concern, we applied our Periodic Focusing Learning Policy (PFLP) that picks a focusing task every 20 iterations and have focusing task's loss weight fixed while multiplying the rest of the task's loss weight by 0.1. In this way, the network for each task gets the opportunity to be dominantly trained. Finally, we proposed the Internal-transfer Weighting (ITW), which "warms up" the network with balanced weights and PFLP, but then applies significant focus (without PFLP) on the central task of cancer detection (Figure 1). In this study, we interrogated the National Lung Screening Trial (NLST) as well as an extensive archive of de-identified screening scans from clinical practice (Vanderbilt Lung Screening Program, VLSP) to evaluate the performance of the proposed method. From the results, the proposed multi-task learning method (Combination of PFLP and ITW) achieves an AUC of the lung cancer detection of 0.8462, which not only outperforms the single-task method (0.8080) by a noticeable margin, but also surpassed other multi-task variants.

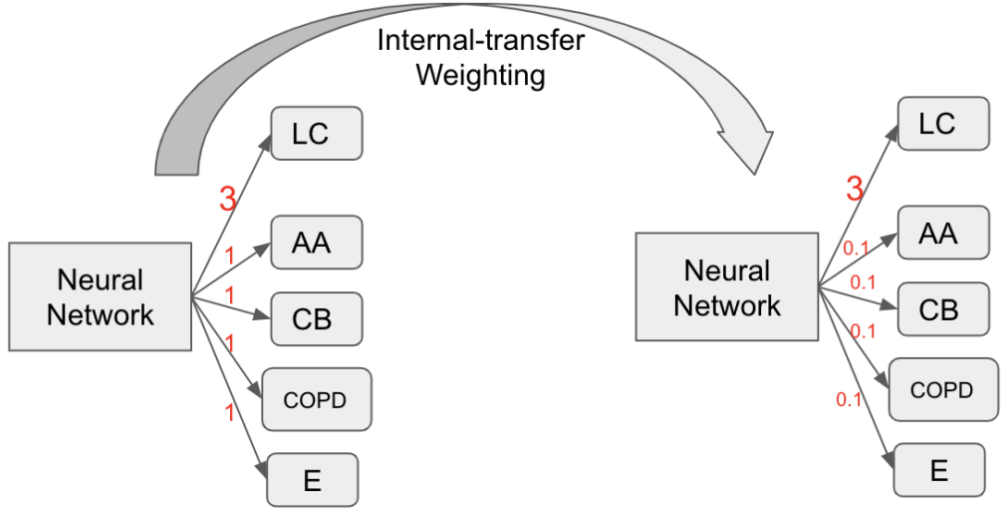

Figure 1. Example of Internal-transfer Weighting (ITW) of multi-task learning. The LC, AA, CB, COPD, E represent the tasks of lung cancer, adult asthma, chronic bronchitis, chronic obstructive pulmonary disease and emphysema, respectively. The red number is the loss weight for the task. Note that we apply PFLP on top of the displayed weights.

## 2. METHODS

### 2.1 Data

We utilized 2512 unique patients' scans and labels from the National Lung Screening Trail (NLST, [3]) dataset as well as 870 unique patients' scans and labels from the de-identified Vanderbilt Lung Screening Program (VLSP) (https://www.vumc.org/radiology/lung) dataset. All data access was in de-identified form under IRB approval. Since NLST is lung cancer centric, reports on other diseases in the study are self-reported. Lacking comprehensive labels is a common limitation in medical imaging datasets, and it could hinder the effectiveness of multi-task in the field. However, using multiple datasets, each with its own accurately labeled task, also suffers from deep learning's inability to generalize across datasets. This drawback of neural networks leads us to shuffle the two datasets together rather than designating one as the training set and the other as the test set. NLST provides self-reports for a set of chronic airway diseases, and we chose the diagnoses of the top four most prevalent kinds of diseases as our auxiliary tasks. We labeled subjects who declined to self-report as not coded. After combining two datasets by random shuffling, we randomly generated a training dataset of 2517 (1904 from NLST, 613 from VLSP) subjects, validation dataset of 277 (192 from NLST, 85 from VLSP) subjects, and a holdout testing set of 592 subjects (420 from NLST, 172 from VLSP). NLST dataset has data for all five tasks, and VLSP has clinical data for lung cancer and COPD, and the remaining three tasks are not coded (NC). Cancer malignancy and diagnosis for the other conditions were well distributed across the three cohorts (Table 1). Because of the inaccuracies in auxiliary tasks' data, we do not anticipate the auxiliary tasks to produce clinical grade predictions, but we expect the auxiliary tasks to give neural network more insights into lung's general wellbeing.

Table 1. Positive Cases for Each Condition During Each Phase

| (NLST/VLSP) | Cancer | Adult Asthma | CB* | COPD* | Emphysema |
|---|---|---|---|---|---|
| Train | 889(876/13) | 120(120/NC*) | 212(212/NC) | 317(212/105) | 209(209/NC) |
| Validation | 54(53/1) | 11(11/NC) | 25(25/NC) | 31(10/21) | 18(18/NC) |
| Test | 132(128/4) | 26(26/NC) | 39(39/NC) | 80(36/44) | 42(42/NC) |
| Total | 1075(1057/18) | 184(184/NC) | 276(276/NC) | 428(258/170) | 269(269/NC) |

An item A(B/C) located at row D column E, means that for condition E during D phase, there are total of A number of positive cases. In those A number of positive cases, B patients are from NLST while C patients are from VLSP. CB – Chronic Bronchitis, COPD – Chronic Obstructive Pulmonary Disease, NC – Not Coded, some labels are not present for all VLSP images. We ignore these outputs from the network during loss calculation.

**2.2 Image Preprocessing**

De-identified data were retrieved from the NLST [3] and VLSP. All images go through the same preprocessing technique, which was developed by Liao et al. [5]. The algorithm accepts images resized to 1mm × 1mm × 1mm isotropic resolution and generates segmented lung image with non-lung area muted with 170 and rest region between [0-255]. The preprocessing technique also generates an estimated nodule mask that we use as the second channel accompanying the main channel (CT scan). We then further downsize the segmented lung CT and nodule mask into 3D float tensor of size 128 × 128 × 128. The final input volume is a 2 × 128 × 128 × 128 4D tensor. The first channel is the image, while the second channel is the nodule mask. Before feeding the image into the network, we perform image augmentation that not only randomly rotates the image on a single axis from 0 to 15 degrees but also randomly translates the 3D image by at most 8 pixels along each of the three main axes.

**2.3 Label Preprocessing**

Our network accepts images with flexible combinations of output labels from the available label pool. This means an image with only one label can still be used in our study since it still provides valuable insight towards that task. Therefore, we selected the most prevalent medical conditions present in NLST dataset and VLSP as labels and selected all images that has at least one of the labels. We used -999 as place holders to indicate "not coded" (see Table 1) for an image's missing label so that the network can ignore this field during loss calculation. Among all the available lung-related labels, we selected 1. Lung Cancer (LC), 2. Adult Asthma (AA), 3. Chronic Bronchitis (CB), 4. Chronic Obstructive Pulmonary Disease (COPD) 5. Emphysema (E) as our tasks. Table 1 delineates the condition break down for each dataset and for each phase (Train / Validation / Testing).

**2.4 Multi-task Neural Network**

The attention-based 3D convolutional network [4] is employed as the backbone network. The details have been provided in the Jiachen et al. [4] paper. As shown in Figure 2, we extended the backbone network by giving each task its independent branch of fully connected layers after global average pooling. This means all tasks only share the encoding part of the network up to the global average pooling layer. The encoding part will provide "task shared" feature maps for the following task-specific learning.

**2.5 Hyper Parameters and Tools**

The project is implemented using Python 3.6 and PyTorch 1.0. With the support of Nvidia's 2080Ti GPU, we were able to feed 4 images into the network per batch. Our learning rate is set to 0.0001 while our backpropagation utilizes Adam algorithm. Furthermore, in order to compare the performance between single-task and multi-task neural network, we took the single-task network as it is and added multi-task functionality on top of it. All the parameters for the single-task neural network were untouched except for the loss weights between classes (Interclass Loss Weights or IW). We changed it from single-tasks' loss weight of only value 3 to multi-task's lists with

$$[3 : w_{AA} : w_{CB} : w_{COPD} : w_E]$$

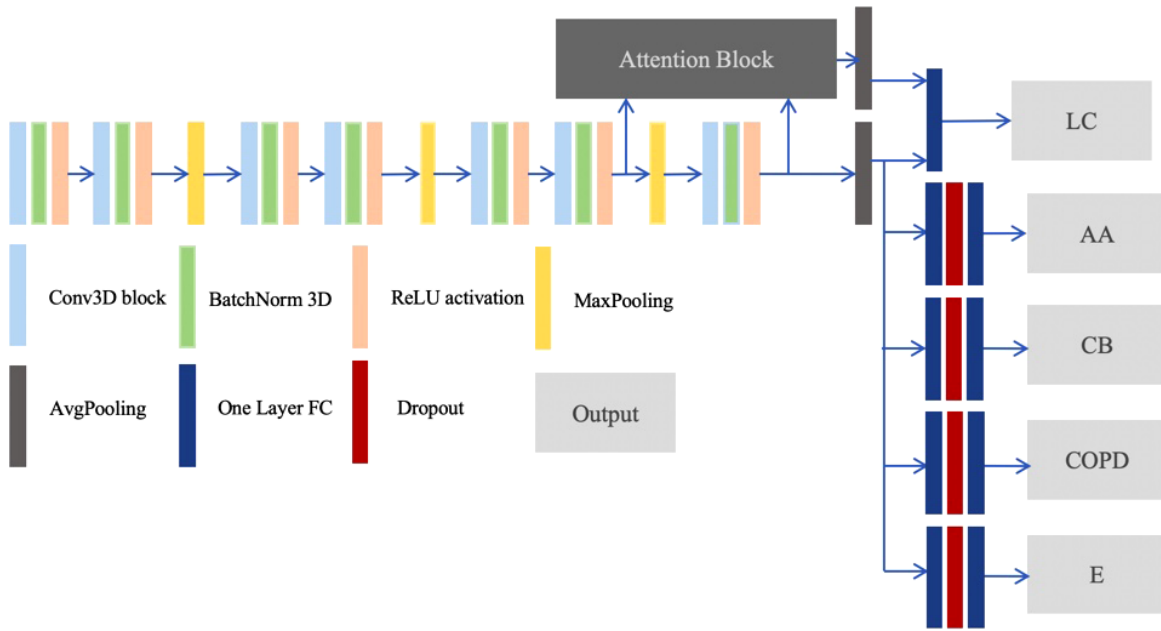

Figure 2. Proposed multi-task neural network. The encoding part of the network is shared by different tasks, while each task has its own subnetwork. Different from Liao et al., which requires lung nodule location to train the nodule detection and classification simultaneously, our method is able to utilize scans without nodule annotation by using a pre-trained nodule detection tool.

Table 2. Multi-task Experiments

|  | No PFLP |  | With PFLP |  |
|---|---|---|---|---|
|  | Name | Loss Weights | Name | Loss Weights |
| No ITW | Multi-task Learning (MTL) | 3: 3: 3: 3: 3 | Multi-task Learning 0 (MTL0) | 3: 0.01: 0.01: 0.01: 0.01 |
|  |  |  | Multi-task Learning 1 (MTL1) | 3: 0.1: 0.1: 0.1: 0.1 |
|  |  |  | Multi-task Learning 2 (MTL2) | 3: 1: 1: 1: 1 |
|  |  |  | Multi-task Learning 3 (MTL3) | 3: 3: 3: 3: 3 |
|  |  |  | Multi-task Learning 4 (MTL4) | 3: 6: 6: 6: 6 |
| With ITW | Internal-transfer Weighting (ITW1) | 3: 3: 3: 3: 3 | Internal-transfer Weighting 2 (ITW2) | 3: 1: 1: 1: 1 |
|  |  |  | Internal-transfer Weighting 3 (ITW3) | 3: 3: 3: 3: 3 |

We ran 9 multi-task experiments and used them to evaluate the effectiveness of our PFLP and ITW weight adapting policies.

### 2.6 Internal-transfer Weighting

When a model's validation set AUC not only reached a peak that is no longer consistently surpassed in the subsequent epochs, but also appears to have plateaued, we assume that the network has learned as much as possible from the other tasks. Then, we used Internal-transfer Weighting which fixed auxiliary tasks' loss weights to 0.1. In turn, our network can focus on improving lung cancer prediction till the end. This resembles a process of "learn comprehensively and then dig in deep" analogy.

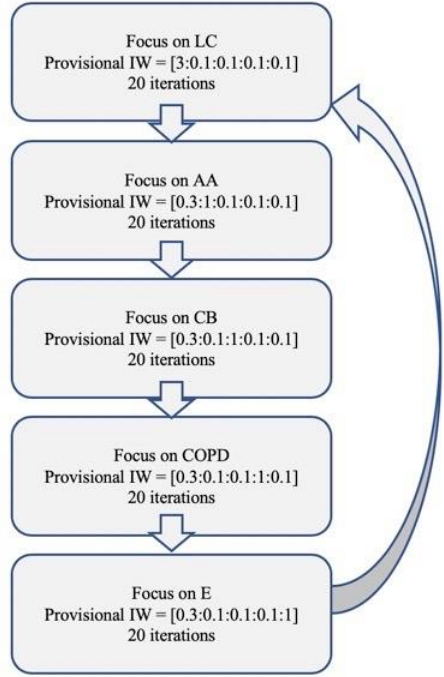

Figure 3. Example of a PFLP loop when initially fixed IW is [3 : 1 : 1 : 1 : 1]. In each case, the focusing task's loss weight remains unchanged but the auxiliary task's loss weights are multiplied by 0.1. The averaged ratio remained to be [3: 1 : 1 : 1 : 1] but each task got the opportunity to be trained without interference from other tasks. If an epoch has X iterations, $\frac{X}{100}$ loops would be performed. A fresh new loop is initiated at the beginning of each epoch with focus on lung cancer.

### 2.7 Periodic Focusing Learning Policy

Initially, no learning took place in multi-task baseline (No PFLP nor ITW) and we noted the difficulty in learning lung cancer and auxiliary airway diseases simultaneously. Therefore, we introduced PFLP that alternates dominance of training among tasks to avoid network confusion. We utilized Periodic Focusing Learning Policy, as shown in Figure 3. We generated provisional inter-class weights based on the inter-class weights in the configuration. We obtain provisional inter-class weight by leaving main task weight fixed and multiplying non-primary task's loss weights by 0.1. The focusing task changed every 20 iterations. When PFLP was applied, the final loss was calculated by weighted sum of losses using the provisional IW. PFLP terminates when we apply ITW to focus exclusively on lung cancer.

### 2.8 Experiments

We ran 3 baselines and 8 comparison experiments. Baselines include Kaggle [5], single-task and multi-task baseline. In all three cases, we did not apply any adaptive weight policy. Table 2 lists the multi-task experiments we performed and the inter-class weights we attempted. In order to compare the effectiveness of PFLP and ITW, we performed Multi-task baseline without adaptive weight policy, multi-task with PFLP, multi-task with ITW and multi-task with PFLP and ITW. We further explored different inter-class weights combinations in the multi-task with PFLP experiments.

### 2.9 Statistical analysis

We utilized the McNemar test, which produces a $\mathcal{X}^2$ statistic based on two predictors' predictions, to validate the improvements.

# 3. RESULTS

The proposed approach towards multi-task lung cancer recognition produced superior results compared with other lung cancer detection approaches as measured by AUC. Table 3 presents every model's best performing epoch (selected based on lung cancer AUC) in complete detail. For models that performed superior to STL, we calculated McNemar p-values. Figure 4 compares each experiment's lung cancer prediction ROC curve, and Figure 5 demonstrates the optimal range of inter-class Weights. First, MTL and ITW1 displayed no signs of learning, i.e., the AUC in training and validation phase oscillated intensely around 0.5. In contrast, MTL1, MTL2 and MTL3 (with PFLP) demonstrate improvement in discrimination from the baseline task with AUC values of 0.8223, 0.8153 and 0.8402 and produce significant ($p < 0.05$) McNemar test p-values with respect to STL. This gives us the confidence to reject the null hypothesis (no improvement from baseline STL). Furthermore, the ITW experiments (both 2 and 3) showed the most noticeable performance lift with AUC values of 0.8462 and 0.8433 on the test set. In comparison, there is no significant difference in performance on the validation set (though both ITW experiments still performed better on validation set), which might be attributed to the validation set's small sample size. Secondly, the auxiliary tasks' prediction performance displayed a general trend of improving as loss weights for auxiliary tasks increases. The degradation for the auxiliary tasks is expected when we changed IW from $[3:3:3:3:3]$(MTL3) to $[3:6:6:6:6]$(MTL4). In this case, auxiliary tasks gained slightly more focus in MTL4 as compared to MTL3 but focus on lung cancer decreased significantly. This matches the significant degradation in MTL4's lung cancer prediction AUC. The auxiliary tasks' performances, compared to that of lung cancer, are not ideal. This is expected because the datasets are lung cancer driven and thus have fewer positive cases for the auxiliary airway diseases.

Furthermore, we observe an optimal range of weights for the multi-task network to perform well further validates our hypothesis. Giving too little focus to other tasks (MTL0), the multi-task learning produced worse results than single-task learning. In contrast, the best performing MTL experiment was the experiment with equal focus (MTL3) and overall best experiment is ITW2 further supports the idea that "study broad and dig in deep" may be the optimal approach for lung cancer detection. Figure 4 shows that MTL0 and MTL4 perform worst among all experiments while MTL3 and both ITW2 and ITW3 surpasses the rest on the test set.

Table 3: Experiment AUC Results

|        | Epoch | LC(Train/Val/Test)        | p-value | AA     | CB     | COPD   | EE     |
| ------ | ----- | ------------------------- | ------- | ------ | ------ | ------ | ------ |
| Kaggle | N/A   | -- / -- / 0.7909          | --      | N/A    | N/A    | N/A    | N/A    |
| STL    | 66    | 0.7705 / 0.8525 / 0.8080  | --      | N/A    | N/A    | N/A    | N/A    |
| MTL    | 6     | 0.5258 / 0.7172 / 0.6720  | --      | 0.6434 | 0.5788 | 0.6759 | 0.6378 |
| MTL0   | 27    | 0.6399 / 0.8051 / 0.6902  | --      | 0.5839 | 0.5067 | 0.5709 | 0.6498 |
| MTL1   | 67    | 0.8013 / 0.8434 / 0.8223  | < 0.01  | 0.5836 | 0.4439 | 0.6642 | 0.5674 |
| MTL2   | 72    | 0.7763 / 0.8368 / 0.8153  | 0.012   | 0.6369 | 0.5839 | 0.7300 | 0.6554 |
| MTL3   | 87    | 0.7975 / 0.8654 / 0.8402  | < 0.01  | 0.6432 | 0.6163 | 0.7592 | 0.6444 |
| MTL4   | 50    | 0.7189 / 0.7865 / 0.7537  | --      | 0.6514 | 0.6012 | 0.7502 | 0.6405 |
| ITW1   | 6     | 0.5258 / 0.7172 / 0.6720  | --      | 0.6434 | 0.5788 | 0.6759 | 0.6378 |
| ITW2   | 97    | 0.8049 / 0.8729 / 0.8462  | < 0.01  | 0.4916 | 0.5148 | 0.7079 | 0.6410 |
| ITW3   | 108   | 0.8130 / 0.8624 / 0.8437  | < 0.01  | 0.6062 | 0.6013 | 0.7641 | 0.6507 |

This tables shows the AUC results from each experiment on each condition. The lung cancer p-values are calculated using McNemar test with respect to STL. The blue values of multitask indicate the lung cancer detection performance better than single-task, and the red value represents the highest performance. Note that data augmentation (e.g., rotation and cropping) during training can cause instability in training set performance, and so the training performance reported here is unstable and less indicative of model performance. It is important to note that both multi-task baseline (MTL) and Internal-transfer Weighting 1 (ITW 1, based on MTL) failed to converge.

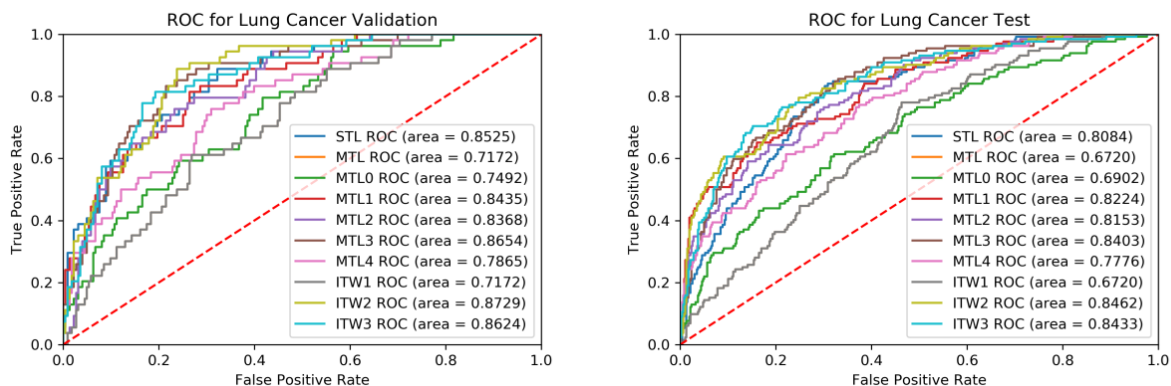

Figure 4. Lung Cancer Validation and Test ROC Curve for all experiments.

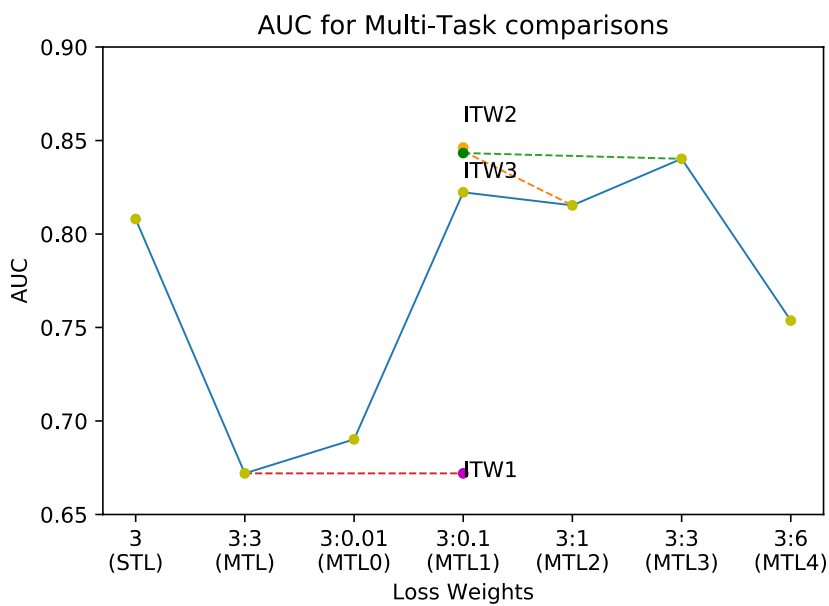

Figure 5. AUC comparisons. The ratio A: B at the x-axis represents the loss weights between the task of lung cancer detection (as A) and the rest auxiliary tasks (as B). Orange line shows ITW2's transition from 3:1 to 3:0.1 and green line shows IT3's transition from 3:3 to 3:0.1. It can be seen in the plot that (1) PFLP helps multi-task networks to converge. (2) When the ratio is appropriate, multi-task experiments surpassed single-task baseline. (3) Internal-transfer Weighting improves multi-task performance.

## 4. CONCLUSION AND DISCUSSION

From the experiments, the PFLP helped the network to converge even though the auxiliary airway disease identification are hard tasks to co-learn. The weights are tuned empirically across different tasks to leverage the performance. In the future, it would be appealing to design the adaptive weight adjustment method in a data-driven manner. Last, based on multi-task learning, ITW further improves the central task's performance by focusing the network exclusively on lung cancer during later stages.

## 5. ACKNOWLEGEMENTS

This research was supported by NSF CAREER 1452485 and NIH R01 EB017230. This research was conducted with the support from Intramural Research Program, National Institute on Aging, NIH. This study was supported in part by a UO1

CA196405 to Massion. This study was in part using the resources of the Advanced Computing Center for Research and Education (ACCRE) at Vanderbilt University, Nashville, TN. This project was supported in part by the National Center for Research Resources, Grant UL1 RR024975-01, and is now at the National Center for Advancing Translational Sciences, Grant 2 UL1 TR000445-06. We gratefully acknowledge the support of NVIDIA Corporation with the donation of the GPU used for this research. The imaging dataset(s) used for the analysis described were obtained from ImageVU, a research repository of medical imaging data and image-related metadata. ImageVU is supported by the VICTR CTSA award (ULTR000445 from NCATS/NIH) and Vanderbilt University Medical Center institutional funding. ImageVU pilot work was also funded by PCORI (contract CDRN-1306-04869). This research was also supported by SPORE in Lung grant (P50 CA058187), University of Colorado SPORE program, and the Vanderbilt-Ingram Cancer Center.

## 6. REFERENCES

[1] N. Khosravan, U. Bagci, "Semi-Supervised Multi-Task Learning for Lung Cancer Diagnosis," 2018 40th Annual International Conference of the IEEE Engineering in Medicine and Biology Society (EMBC), Honolulu, HI, 2018, pp. 710-713 (2018)

[2] S. Hussein, K. Cao, Q. Song, U. Bagci "Risk Stratification of Lung Nodules Using 3D CNN-Based Multi-task Learning," Information Processing in Medical Imaging. IPMI 2017. Lecture Notes in Computer Science vol 10265, (2017)

[3] N. L. S. T. R. Team, "The national lung screening trial: overview and study design," Radiology 258(1), 243-253 (2011).

[4] J. Wang, R. Gao, Y. Huo, et al., "Lung cancer detection using co-learning from chest CT images and clinical demographics," Proc. SPIE 10949 Medical Imaging 2019: Image Processing, (2019)

[5] F. Liao, M. Liang, Z. Li et al., "Evaluate the Malignancy of Pulmonary Nodules Using the 3D Deep Leaky Noisy-or Network," arXiv, (2017).

[6] Q. Liao, L. Jiang, X. Wang, et al., "Cancer classification with multi-task deep learning," 2017 International Conference on Security Pattern Analysis and Cybernetics (SPAC) , pp. 76-81 (2017)

[7] X. Zhu, J. Yao and J. Huang, "Deep convolutional neural network for survival analysis with pathological images," 2016 IEEE International Conference on Bioinformatics and Biomedicine (BIBM) , pp. 544-547 (2016)

[8] "Cancer," World Health Organization, [Online]. Available: https://www.who.int/news-room/fact sheets/detail/cancer. (2018)

[9] "Final Update Summary: Lung Cancer: Screening," U.S. Preventive Services Task Force. https://www.uspreventiveservicestaskforce.org/Page/Document/UpdateSummaryFinal/lung-cancer-screening, (2015)

[10] R. Gao, Y. Huo, S. Bao, et al., "Distanced LSTM: Time-Distanced Gates in Long Short-Term Memory Models for Lung Cancer Detection," (2019)

[11] D. Ardila, A.P. Kiraly, S. Bharadwaj, et al., "End-to-end lung cancer screening with three-dimensional deep learning on low-dose chest computed tomography," Nature Medicine 25, 954–961 (2019)

[12] L. Liu, Q. Dou, H. Chen et al. "MTMR-Net: Multi-task Deep Learning with Margin Ranking Loss for Lung Nodule Analysis," Deep Learning in Medical Image Analysis and Multimodal Learning for Clinical Decision Support, 74-82 (2018)